\documentclass{Interspeech}

\interspeechcameraready

\title{Evaluating Speech-to-Text Systems with PennSound}

\author{Jonathan}{Wright}
\author{Mark}{Liberman}
\author{Neville}{Ryant}
\author{James}{Fiumara}

\affiliation{Linguistic Data Consortium}{University of Pennsylvania}{USA}
\email{jdwright@upenn.edu, myl@upenn.edu, nryant@upenn.edu, jfiumara@upenn.edu}

\usepackage{comment}
\usepackage[inkscapelatex=false]{svg}
\usepackage{kotex}

\begin{document}

\maketitle

\begin{abstract}
    

A random sample of nearly 10 hours of speech from PennSound, the world's largest online collection of poetry readings and discussions, was used as a benchmark to evaluate several commercial and open-source speech-to-text systems. PennSound's wide variation in recording conditions and speech styles makes it a good representative for many other untranscribed audio collections. Reference transcripts were created by trained annotators, and system transcripts were produced from AWS, Azure, Google, IBM, NeMo, Rev.ai, Whisper, and Whisper.cpp. Based on word error rate, Rev.ai was the top performer, and Whisper was the top open source performer (as long as hallucinations were avoided). AWS had the best diarization error rates among three systems. However, WER and DER differences  were slim, and various tradeoffs may motivate choosing different systems for different end users. We also examine the issue of hallucinations in Whisper. Users of Whisper should be cautioned to be aware of runtime options, and whether the speed vs accuracy trade off is acceptable.

\end{abstract}

\section{Introduction}

PennSound\footnote{\url{https://writing.upenn.edu/pennsound/}} is the world's largest online collection of poetry readings and discussions\footnote{Charles Bernstein, co-creator, \\  \url{https://www.youtube.com/watch?v=mClvCzCOHaE}}. The recordings are free to download for non-commercial purposes.   This work is part of an effort to evaluate the efficacy of current speech-to-text technology in indexing this collection. The varied nature of the PennSound collection, which includes interviews and discussions as well as readings, means that this evaluation may also be relevant  to other large-scale audio datasets.  

We chose a random sample of 100 five-minute or longer selections\footnote{See supplementary materials for details.} from this data to evaluate several leading speech-to-text systems.  The sample includes a broad range of conditions, including varying recording quality, number of speakers, and speaking style.

\section{Methods}

\begin{figure}[t]
\centering
\includesvg[width=\linewidth]{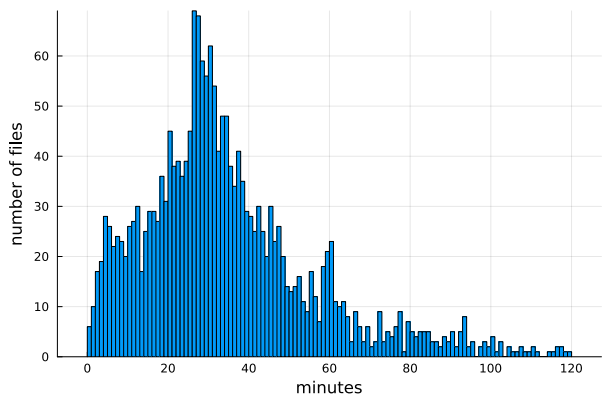}
\caption{Distribution of full length recording durations.}
\label{fig:hist1}
\end{figure}

The PennSound archive includes over 30,000 links to mp3s, although many of these files are individual poems excerpted from larger recordings.  Over two thousand files downloaded from the site averaged 37.9 minutes in duration, for a total of 1,295 hours, out of a total of more than 6,000 hours.  The histogram in Figure \ref{fig:hist1} shows the distribution of durations for this subset of the collection.

A random selection of 100 sources was made from 1,869 .mp3 URLs on the PennSound website that fall under the \emph{Authors} page, and have the word \emph{Complete} in the file name, indicating a complete recording.  Two files outside this list, from \emph{Close Listening} and \emph{Poem Talks}, were arbitrarily added to ensure examples with multiple speakers.

  Speech activity labels were produced using an HMM-based speech activity detector (SAD) trained on English interviews
   (\textit{ldc-bpcsad} \cite{sad}) .  Starting from the beginning of a speech segment near the midpoint of each full recording, a stretch of audio was extracted such that at least 5 minutes of speech was included, based on the SAD labels.  These 100 clips became the data for the experiments described below. The sample includes introductions, readings, question-and-answer sessions, and discussions.  Reference transcripts were created by a team of trained annotators using a set of transcription guidelines\footnote{See supplementary materials.} and an online transcription platform 
   \cite{Wright2021-jt}
   .  Several speech-to-text systems were used to generate hypothesis transcripts, including five commercial cloud services:  Amazon Transcribe\footnote{\url{https://aws.amazon.com/transcribe/}} (aws), Microsoft Azure AI Speech\footnote{\url{https://azure.microsoft.com/en-us/products/ai-services/ai-speech/}} (azure),  Google Cloud Speech-to-Text\footnote{\url{https://cloud.google.com/speech-to-text?hl=en}} (google), IBM Watson Speech to Text\footnote{\url{https://www.ibm.com/products/speech-to-text}} (ibm), and Rev.ai\footnote{\url{https://www.rev.ai/async}}  (rev).  NVIDIA's NeMo\footnote{\url{https://github.com/NVIDIA/NeMo}} (nemo), Open AI's Whisper\footnote{\url{https://github.com/openai/whisper}} (whisper), and the related Whisper CPP \footnote{\url{https://github.com/ggerganov/whisper.cpp}} (whispercpp), three open source systems, were also run locally.

The mp3s were converted to \SI{16}{\kilo\hertz} mono using SoX\footnote{\url{https://sourceforge.net/projects/sox/}}, using the \verb|--norm=-1| option to normalize the levels.  Files were then fed to the systems as wav or flac depending on requirements.  Parameter values used with the systems are provided in the supplementary material. The corresponding parameters are described in the documentation for aws\footnote{\url{https://docs.aws.amazon.com/transcribe/latest/APIReference/API\_StartTranscriptionJob.html}}, azure\footnote{\url{https://learn.microsoft.com/en-us/dotnet/api/microsoft.cognitiveservices.speech.speechconfig?view=azure-dotnet}}, google\footnote{\url{https://cloud.google.com/speech-to-text/v2/docs/transcription-model}}, ibm\footnote{\url{https://cloud.ibm.com/docs/speech-to-text?topic=speech-to-text-models-use}},  nemo\footnote{\url{https://docs.nvidia.com/nemo-framework/user-guide/24.09/nemotoolkit/asr/intro.html}},
rev\footnote{\url{https://docs.rev.ai/api/asynchronous/reference/\#operation/SubmitTranscriptionJob}},
whisper\footnote{\url{https://github.com/openai/whisper}}, 
whispercpp\footnote{\url{https://github.com/ggerganov/whisper.cpp}}.

Following previous work \cite{Bain2023-gh,Ferraro2023-uu,Hannun2014-qt,Ravanelli2021-pz}, Word Error Rate (WER) is used to evaluate word recognition by the systems.
WERs were calculated with NIST's Scoring Toolkit (SCTK)\footnote{\url{https://github.com/usnistgov/SCTK}}.  The sctk command calculates the edit distance between reference transcripts and hypothesis transcripts, and the normalized edit distance is the WER.  We also use hubscr.pl from SCTK, which calls sctk after normalizing spelling.  Scoring in this way also involves comparing timestamps, which are not of primary interest here (see Future Work), so the reference transcripts, which have segment boundaries, are converted into single segment transcripts.  As three cloud services\footnote{The chosen settings for google and ibm don't support diarization.} also produce speaker labels, we additionally report diarization error rate (DER) \cite{fiscus2006rich}, calculated using the \textit{dscore} tool \cite{dscore} used by the DIHARD challenges \cite{ryant21_interspeech}. 

\section{Results}

\subsection{Word Recognition}

Word error rates from SCTK for the full sample are shown in Table \ref{table:1}.

\begin{table}[th]
  \caption{Word Error Rates}
  \label{table:1}
  \centering
  \begin{tabular}{l | c c }
    \toprule
    System & WER \\
    \midrule
                        aws & 9.8  \\
                        azure & 10.2 \\
                        google & 11.0 \\
                        ibm & 14.4 \\
                        nemo & 10.9 \\
                        rev & 9.0 \\
                        whisper & 9.5 \\
                        whispercpp & 10.9 \\

    \bottomrule
  \end{tabular}

\end{table}

Other work suggests these WERs are better than what is necessary for effective information retrieval \cite{Hauptmann2002-dy}.

\begin{figure}[t!]
\centering
\includesvg[width=\linewidth]{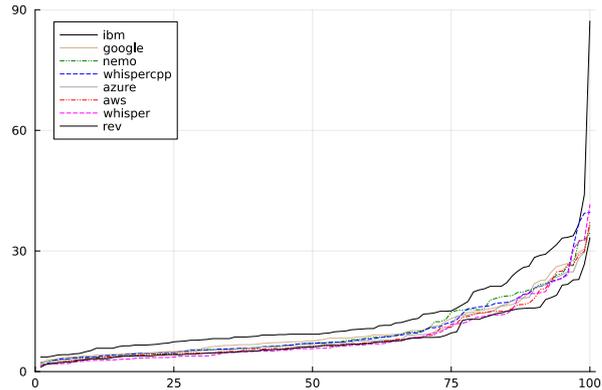}
\caption{WERs, Sorted Separately}
\label{figure:wer1}
\end{figure}

As mentioned earlier, the nature of the files varies quite a bit, from clear readings, to difficult to understand discussions, and while the sample is random, the distribution of these variables is not balanced.  The total WER may obscure performance details, and individual WERs per file provide detail on how the systems vary across files.  Figure \ref{figure:wer1} plots all the individual WERs; each system's WERs have been sorted separately. The complete table of WERs by sample and system is available in the Supplementary Materials.

\subsection{Diarization}

Three systems, aws, azure, and rev, provide speaker labels, which we can evaluate with the dscore package;  DER scores are shown in Table \ref{table:4}.

\begin{table}[th]
  \caption{DERs}
  \label{table:4}
  \centering
  \begin{tabular}{l | r r }
    \toprule

    System & DER \\
    \midrule
                        aws & 14.52  \\
                        azure & 15.40 \\
                        rev & 15.49  \\

    \bottomrule
  \end{tabular}

\end{table}

\begin{figure}[t!]
\centering
\includesvg[width=\linewidth]{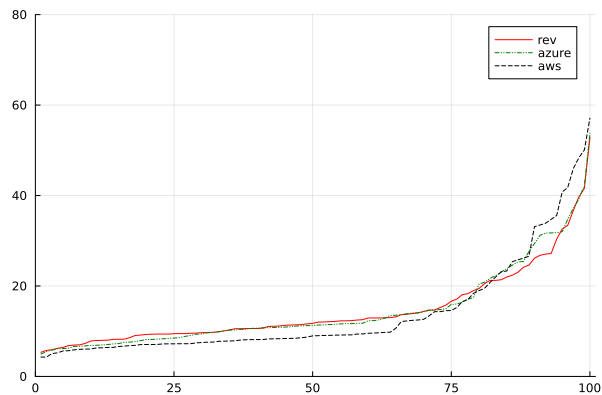}
\caption{DERs, Sorted Separately}
\label{figure:der1}
\end{figure}

As with WERs, we can look at performance by recording.  Figure \ref{figure:der1} plots the DERs for individual recordings, with each system sorted separately.

\section{Discussion}

Although ibm lags slightly, the performance of the systems based on WER alone is effectively the same for end users.  There is a slight tendency for Whisper to do best on the easier recordings, and a slight tendency for Rev.ai to do best on the harder recordings.  AWS, Azure, and Rev.ai are effectively the same based on DER as well.

Whisper is an attractive system, but only if hallucinations are avoided.  Whisper is known to hallucinate, which means that it adds text in silent regions and after the end of the input, as if the transcribed text were a prompt. The inserted text can be harmful as well as imaginary \cite{Koenecke2024-rm}.  One way to characterize the output is to count these tokens that appear beyond the end of the audio, what we call the Final Insertion Count (FIC).  Another way is to look at the Insertion Error Rate (IER), the component of the WER due to insertions.  Table \ref{tab:example} shows the WER, IER, and FIC, for runs of Whisper using different models.  OpenAI provides several models to deal with the trade-off between model memory size and performance.

\begin{table}[th]
  \caption{Audio final hallucinations}
  \label{tab:example}
  \centering
  \begin{tabular}{r l  r r }
    \toprule
    Model & WER & IER & FIC \\
    \midrule
                        tiny  & 17.7 & 2.6 & 89 \\
                        base & 14.3 & 1.7 & 129 \\
                        small & 11.5 & 1.1 & 27 \\
                        medium & 11.1 & 1.1 & 55 \\
                        large & 14.8 & 5.2 & 263 \\
                        large-v2 & 10.5 & 1.0 & 76 \\
                        large-v3 & 15.4 & 5.6 & 124 \\
                        large-v3-turbo & 11.2 & 2.9 & 1601 \\
    \bottomrule
  \end{tabular}
  
\end{table}

The turbo (aka large-v3-turbo) model is a reduction in size of the large-v3 model that should have equivalent performance; the turbo model is the default model and the one used in our main trial.  However, in the main trial, Whisper had a WER of 9.5, an IER of 1.3, and no extraneous tokens beyond audio's end.  This is due to the options OpenAI has recently added to ameliorate the problem of hallucinations\footnote{https://github.com/openai/whisper/pull/1838}:

    \begin{verbatim}
--word_timestamps
--hallucination_silence_threshold
    \end{verbatim}

The \verb|word_timestamps| option must be set to \verb|True|.  We always used either \verb|0.1| or \verb|0.5| for \verb|hallucination_silence_threshold| for the reported results, although the precise value had no obvious effect.

The trials in Table \ref{tab:example} do not use these options.  In the main trial, the average IER of all systems was 1.3, so the larger IERs in Table \ref{tab:example} suggest significant hallucinations (at the end or otherwise).  Also note that the fluctuation in performance across models could be due to the non-deterministic nature of Whisper:  two runs with the same parameters generally yield slightly different transcripts and different hallucinations.  An example of a final hallucination follows:

    \begin{verbatim}
That means, unlike what you
are moving forward in the time, it is it a
little bit wyskidotry decomposition.
That is what a loaded bread
there, that is something like it.
It's going to be better to touch the
electronics barge. That is what it is.
So, here we go. But there,
there are various areas where you're
looking at, the way that is Вс쟁.        
    \end{verbatim}

Several possibilities for the variation in WER are explored in the supplementary materials.  Signal-to-Noise-Ratio as reported by ibm does not appear explanatory, even for ibm; we plot WER by SNR just for ibm in Figure \ref{figure:snr}.

\begin{figure}[t!]
\centering
\includesvg[width=\linewidth]{ibmsnr.svg}
\caption{IBM SNR}
\label{figure:snr}
\end{figure}

\begin{figure}[t!]
\centering
\includesvg[width=\linewidth]{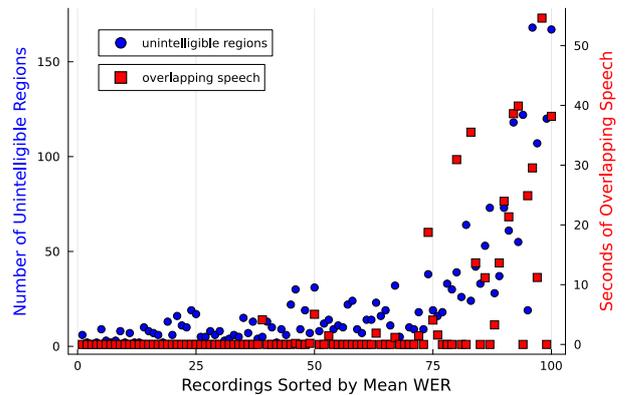}
\caption{Unintelligible regions and overlapping speech explain some high WERs}
\label{figure:unov}
\end{figure}

System WERs often vary sharply but in concert, suggesting some common cause.  One possibility is overlapping speech, where systems tend to capture a single speaker.  Another possibility is the presence of unintelligible regions in otherwise clear recordings of speech.  Transcribers mark such regions, so we can quantify this effect.  In Figure \ref{figure:unov} we plot the amount of overlapping speech in seconds along with the number of marked unintelligible regions (note the differing units) on the y-axis, with the mean WER on the x-axis.

Here's a difficult example where two speakers alternate during a reading as well as overlap with each other.  First is the reference transcript, second is the ibm output, third is the rev output.  For the entire recording, ibm and rev had WERs of 30.4 and 21.5.  Words in all caps are substituion or insertion errors, while asterisks are deletion errors.

\begin{verbatim}
sp1 Brazil is caipirinha with feijoada,
sp1 caipira with fedora
sp2 Brazil is home of #
sp2 I don't know ((that one.))
sp1 the cassava or tapioca,
sp2 Or
sp1 what you call yucca
sp2 ((oh))
sp1 or mandioca or aipim or mogo
sp1 or macaxeira or singkong or tugi
    or balinghoy or manioc
sp2 Brazil is the b- #
sp1 You can probably say most of those
    better than me. Why don't you redo
    that one?
sp2 ((Sure))
sp2 ((Sure, yes.)) No, you're= you're #
    you are great in Portuguese. {laugh}

    
\end{verbatim}

\begin{verbatim}
ibm
brazil is CAPRINO with feijoada CAPRA
with fedora brazil is home of i do not
know * * the SAVA or tapioca * what you
call * JACK or MAN DOC or A PIN or MUG
or MAX SYRIA or * SING KONG or TUG or
* * * * * * * * * * * * * * * * * * *
* * BOWLING JOY OR MANY YEAH you are
you are you are great in *
\end{verbatim}

\begin{verbatim}
rev
brazil is CAPIRINA with feijoada CAPIRA
with fedora brazil is home of i do not
know * * the cassava or tapioca * what
you call yucca * or mandioca or APIN or
MUGU or MAXERIA or SINCANG or TUGUI or
BALINGHOI or * * * * * MANIAC COULD
probably say most of those THAT OTHER
than me why do not * redo that * * * *
* * LAW WELL YEAH you are great in *
\end{verbatim}

An easier example follows.  In the (illustrated start of) this example, rev, aws, google, and whisper scored perfectly, with a WER below 4\% for the whole six-minute sample; ibm only makes two errors in this segment, substituting \emph{charley} for \emph{shyly} and \emph{wheels} for \emph{wails}, with a WER of 10.1 on the whole file; and azure makes one minor error, substituting \emph{dolls} for \emph{doll}.

\begin{verbatim}
Cries of love and alarm on the
soundtrack fade into an air raid siren,
factory whistle, or is it the whistle
of a train approaching? We are as real
and as near as cinema
A little girl half turned away, holding
her doll, smiles shyly. When the
ambiguous siren or engine wails its
warning, she turns directly towards
the camera. Medium closeup, her
expression changing to a mixture of
astonishment or terror. This child
isn't acting
\end{verbatim}

\section{Future Work}

This experiment on a representative PennSound sample demonstrates that several different speech-to-text systems have low enough error rates that they will probably provide a useful basis for text-based document retrieval. Our next step is therefore to create automatic transcriptions of the full collection, and to implement a web-based front end for searching and exploring both text and audio of the corpus.  We also will continue to use this sample as a source of benchmark results for future improvements in speech-to-text systems.

We will also evaluate time-mediated evaluation of word recognition.  In the current sample, the reference transcripts were converted into single segments before scoring, in order to eliminate insertion and deletion errors caused by the correct word occurring in the wrong time segment.  This level of precision will be relevant in some situations.  Evaluation will require greater scrutiny of the reference timestamps that serve as ground truth.

\section{Appendix}

The audio sample is available at

    \url{https://github.com/Linguistic-Data-Consortium/penn_sound_audio}

Supplementary Material (transcripts, code, tables, etc.) is available at

    \url{https://github.com/Linguistic-Data-Consortium/penn_sound_eval}

\section{Acknowledgements}

We are grateful to Charles Bernstein, Chris Mustazza, and Al Filries at PennSound for their collaboration on this project.

\bibliographystyle{IEEEtran}
\bibliography{mybib}

\end{document}